\documentclass{article}

\usepackage{arxiv}

\usepackage[utf8]{inputenc} % allow utf-8 input
\usepackage[T1]{fontenc}    % use 8-bit T1 fonts
\usepackage{hyperref}       % hyperlinks
\usepackage{url}            % simple URL typesetting
\usepackage{booktabs}       % professional-quality tables
\usepackage{amsfonts}       % blackboard math symbols
\usepackage{nicefrac}       % compact symbols for 1/2, etc.
\usepackage{microtype}      % microtypography
\usepackage{lipsum}
\usepackage{graphicx}
\graphicspath{ {./images/} }
\usepackage{paracol}
\usepackage{graphicx}
\usepackage[utf8]{inputenc}
\usepackage{caption}
\usepackage{amssymb}
\usepackage{amsmath}
\usepackage{algorithm}
\usepackage[noend]{algorithmic}
\usepackage{cite}
\usepackage{float}
\usepackage{caption,subcaption}
\usepackage{lineno}

\usepackage{siunitx}
\usepackage{booktabs}
\usepackage{mathtools}

\usepackage{pifont}% http://ctan.org/pkg/pifont
\usepackage{url}

\usepackage{xspace}%
\newcommand{\etal}{\textit{et al.\xspace}}

\title{Automated Assignment Grading with Large Language Models: Insights From a Bioinformatics Course}

\author{
    Pavlin G. Poli\v{c}ar \\
    University of Ljubljana\\
    Faculty of Computer and Information Science\\
    Večna pot 113, 1000 Ljubljana, Slovenia\\
    \texttt{pavlin.policar@fri.uni-lj.si} \\
\And
    Martin \v{S}pendl \\
    University of Ljubljana\\
    Faculty of Computer and Information Science\\
    Večna pot 113, 1000 Ljubljana, Slovenia\\
\And
Toma\v{z} Curk \\
    University of Ljubljana\\
    Faculty of Computer and Information Science\\
    Večna pot 113, 1000 Ljubljana, Slovenia\\
\And
    Bla\v{z} Zupan \\
    University of Ljubljana\\
    Faculty of Computer and Information Science\\
    Večna pot 113, 1000 Ljubljana, Slovenia\\
    \\
    Department of Education, Innovation and Technology\\
    Baylor College of Medicine\\
    1 Baylor Plaza, TX 77030, USA
}

\date{}

\begin{document}
\maketitle

\begin{abstract}
\textbf{Motivation:}
Providing students with individualized feedback through assignments is a cornerstone of education that supports their learning and development. Studies have shown that timely, high-quality feedback plays a critical role in improving learning outcomes. However, providing personalized feedback on a large scale in classes with large numbers of students is often impractical due to the significant time and effort required. Recent advances in natural language processing and large language models (LLMs) offer a promising solution by enabling the efficient delivery of personalized feedback. These technologies can reduce the workload of course staff while improving student satisfaction and learning outcomes. Their successful implementation, however, requires thorough evaluation and validation in real classrooms.\\
\textbf{Results:}
We present the results of a practical evaluation of LLM-based graders for written assignments in the 2024/25 iteration of the Introduction to Bioinformatics course at the University of Ljubljana. Over the course of the semester, more than 100 students answered 36 text-based questions, most of which were automatically graded using LLMs. In a blind study, students received feedback from both LLMs and human teaching assistants without knowing the source, and later rated the quality of the feedback. We conducted a systematic evaluation of six commercial and open-source LLMs and compared their grading performance with human teaching assistants. Our results show that with well-designed prompts, LLMs can achieve grading accuracy and feedback quality comparable to human graders. Our results also suggest that open-source LLMs perform as well as commercial LLMs, allowing schools to implement their own grading systems while maintaining privacy.
\end{abstract}


\section{Introduction}

The recent development and widespread availability of large language models (LLMs) have led to their adoption across numerous fields of human endeavor~\cite{kaddour2023, minaee2024}. Their ability to provide instant and personalized responses has naturally prompted researchers to explore their use in education, revealing applications that benefit both students and instructors. These applications take various forms, including personalized student tutoring~\cite{Lyu2024}, contextualizing exercises to enhance engagement~\cite{Yadav2023}, and automated grading of student submissions~\cite{Chiang2023, Liu2023}.

In addition to reducing the workload on teaching faculty, automated grading offers numerous benefits to students and their educational outcomes. Studies have shown that students prefer feedback that is both linguistically clear and provided in a timely manner~\cite{Paterson2020}. Encouraging and constructive feedback has also been linked to improved academic performance. Furthermore, automated grading ensures greater consistency in scoring and feedback, as LLMs are not prone to human errors such as fatigue and variability in grading standards~\cite{KLEIN20021023, madigan_teacher_2023}. This approach allows teaching assistants to dedicate more time to direct interactions with students, which students also highly value~\cite{Paterson2020}.

Automatic grading of student assignments dates back to as early as 1968~\cite{Page1968}. Since then, several systems for grading short answers have been developed, typically relying on a corpus of annotated responses~\cite{Mohler2011, Riordan2017}. However, the emergence of LLMs with few-shot capabilities makes them particularly well-suited for automated grading, especially in cases where instructors can anticipate correct answers and common mistakes. As a result, adopting this technology has become more feasible than ever.

Several studies have explored the use of LLMs in the classroom. Kostić \etal~\cite{Kostic2024} examined GPT-4's performance in essay grading and found that it performed poorly. They also investigated grading variability among three human instructors in a small workshop setting; however, their study was limited to only three instructors grading four essays. Chiang \etal~\cite{Chiang2024} integrated GPT-4 into a real-world course, Introduction to Generative AI, where students had access to the LLM and grading prompts, allowing them to test their answers up to 80 times per assignment. In this case, students' final grades were determined by the scores they were able to achieve using the LLM. Impey \etal~\cite{Impey2024} applied GPT-4 to grade submissions from three massive open online courses, where it outperformed peer grading. However, their approach relied solely on correct answers and grading rubrics in the prompts, focusing on assigning the best possible grade while overlooking the importance of providing constructive feedback. Dai \etal~\cite{Dai2023}, on the other hand, successfully used ChatGPT to generate feedback for student submissions.

%\iffalse
\begin{figure}[htb]
    \centering
    \includegraphics[width=0.5\linewidth]{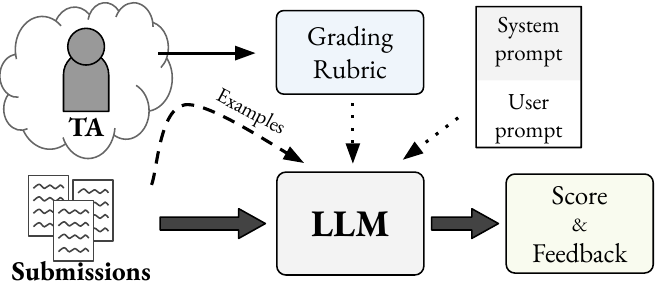}
    \caption{Schema of student submissions graded by LLMs, based on TA-graded examples and grading rubric composed of criteria.}
    \label{fig:LLM-grading-schema}
\end{figure}
%\fi

In this study, we examine the use of LLM graders in a university classroom setting applied to the Introduction to Bioinformatics course, a hands-on bioinformatics course whose innovative design and focus on practical problems we previously reported at ISMB-24~\cite{Policar2024}. Unlike Chiang \etal~\cite{Chiang2024}, where students had access to LLM-generated grading prompts, we utilized LLMs as direct replacements for human graders, grading student submissions only once after the assignment due date, without providing students access to the grading prompts (see Fig.~\ref{fig:LLM-grading-schema}). This setup closely reflects real-world grading scenarios and serves as a valuable case study for implementing LLMs in other academic settings. Additionally, the study was conducted in a randomized manner, where students were unaware of whether their submissions were graded by a human or an LLM. Students subsequently evaluated the quality of the feedback they received, enabling a quantitative comparison between human and machine grading. While most existing studies focus on a single LLM, typically GPT-4, we systematically compare the performance of six different LLMs as automated graders and benchmark them against human teaching assistants.

The study design was reviewed and approved by the Research Ethics and Data Handling Review Board of the University of Ljubljana (approval number 20241130001) to ensure compliance with ethical research standards.

\section{Study Design}
\label{sec:study_design}

We conducted our study in the introductory course to bioinformatics offered by the Faculty of Computer and Information Science, University of Ljubljana, during the 2024/25 winter semester. The course is taught in English.
This year's cohort included 119 students, primarily master's level computer science students, but also included several students from the Faculty of Mathematics and Physics and the Biotechnical Faculty.
The course comprises lectures, five take-home assignments, and a final exam. Each of the five take-home assignments tackles a different aspect of bioinformatics, following the SARS-CoV-2 case study detailed in our previous work~\cite{Policar2024}.
Each assignment consists of multiple exercises in which students implement bioinformatics algorithms, apply them to real-world data, visualize their findings, and discuss their results in written answers to specific questions.
Each assignment contains several mandatory exercises designed to guide students through an investigation of the SARS-CoV-2 virus.
Students can earn extra points by completing bonus exercises that complement the main storyline.
After each assignment deadline, the TAs assess each student's submission and provide a numeric score for the overall assignment and written feedback clarifying mistakes and offering potential improvements.
In our standard execution of the course, programming exercises are graded using automatic unit tests, while figure submissions and text answers are graded manually by the TA.

In the present study, we investigate whether LLMs could be used in place of human TAs for the assessment and grading of written text answers.
Participating students had their text-based answers reviewed and graded by an LLM. Unless a human review of the grade was requested by the student, the LLM-assigned grades were used in their final grades.
Consent was obtained for each of the five assignments.
Participation was purely voluntary, and a student's decision on whether or not to participate had no bearing on the student's final grades.
Students withholding their consent had their assignments graded in our standard manner, using automated unit tests and human review.
Study participation rates were high. On average, we received 105 submissions for each of the five assignments, where between 99 and 101 ($\sim 94$\%) students gave consent to be included in the study. Overall, 93 students gave consent for all five assignments.

%In our study, participating students had their text answers graded by one of eight different graders, including human TAs and LLMs.
The study was performed as follows.
Each of the five assignments includes between 2 to 7 mandatory essay-style questions and between 1 to 3 optional bonus essay-style questions.
Each textual response was randomly assigned to one of the eight graders, that, based on predefined criteria, assigned a score and provided written feedback on the student submission. This feedback was interspersed with unit test-generated feedback from programming exercises and TA-written feedback for figure submissions.
Consequently, students receive grades and feedback from multiple graders on textual questions in a single assignment. The students are not informed by which grader was assigned to each text-based answer.
Students do not have access to the prompts at any point.
Upon receiving their assignment grade and feedback, we ask students to fill out a survey rating their satisfaction with the feedback on each of the text-based questions in their assignment.
Due to the potential for LLM errors, participating students may request a human review of any of the answers. If no re-evaluations are requested, the LLM-assigned grades are used as their final grades.

To assess the capabilities of LLMs for grading student-written free-text submissions, we include three different LLM model architectures, including the popular ChatGPT model (GPT-4o) from OpenAI~\cite{Hurst2024}, four different versions of the open-source Llama 3 models from Facebook~\cite{llama3modelcard}, and a recent model from NVIDIA (Llama-3.1-Nemotron-70B, referred to as Nvidia-70B)~\cite{Wang2024}. Facebook released three open-source versions of the Llama 3 architecture with varying numbers of parameters: 7B, 70B, and 405B. While the larger of these models require specialized hardware which is often not available to university departments, the smaller models can be run on high-end consumer-grade GPUs, which can more readily be found in university departments. Additionally, the hardware requirements can often be reduced through quantization, often at minimal loss in performance~\cite{Jin2024}.
In our study, we include the full-precision versions of Llama-8B and Llama-70B, as well as quantized versions of Llama-70B and Llama-405B, which we subsequently denote as Llama-70Bq4 and Llama-405Bq4, respectively. The full-precision version of Llama-405 was not included due to hardware limitations, while a quantized version of Llama-8B was not included based on poor performance in preliminary preparations for this study.
In total, we include six LLMs: GPT-4o, Nvidia-70B, Llama-405Bq4, Llama-70B, Llama-70Bq4, and Llama-8B.

One of the key things an LLM grader must be able to do is provide good feedback to the students. As previously described, we establish the quality of the feedback using user surveys that students fill out after receiving their feedback. However, there are multiple aspects that humans take into account when evaluating written feedback, of which we identify tone and content as the two most important aspects.
To help disentangle the degree to which students prefer the tone of LLM responses to the content of LLM responses, we include an additional grading group that includes human TA-written feedback, corrected with LLMs tone of writing.
In this grading group, human TAs provide scores and written feedback for each assignment. This feedback is then revised by GPT-4o-mini, where the prompt contains instructions not to alter the content of the feedback but merely rewrite the prompt in the typical ChatGPT style.
We refer to this grading group as ``TA-GPT-revised''.
This way, if any student feedback preferences are established, we can determine whether differences occur merely due to the tone of the feedback or also due to the content of the feedback.

\section{Prompts}
\label{sec:prompts}

\begin{figure*}[htp]
    \centering
    \includegraphics[width=\linewidth]{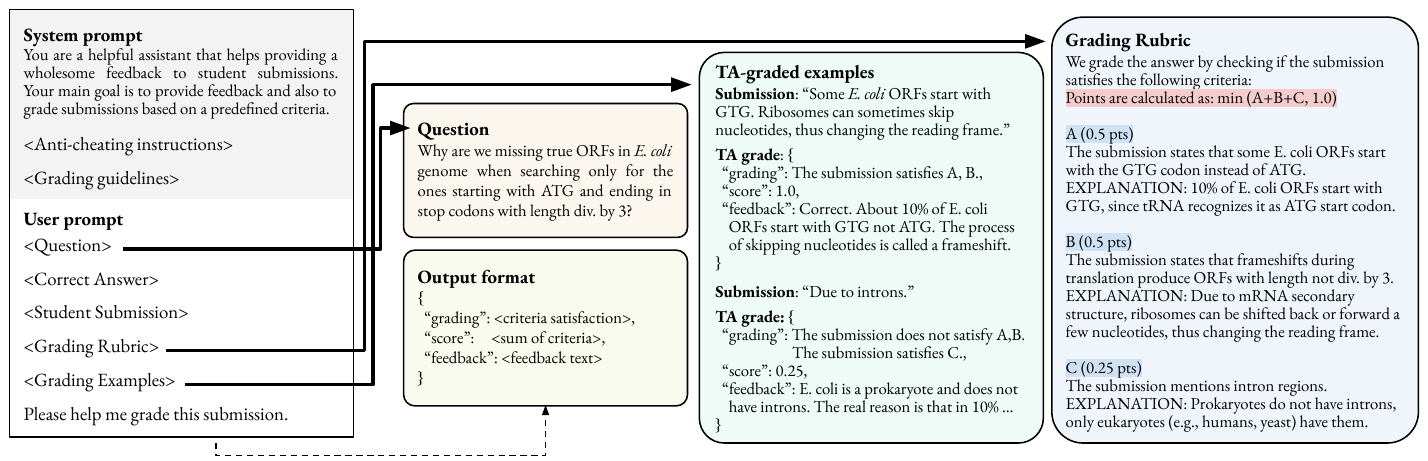}
    \caption{Prompt structure with a grading rubric and TA-graded examples. The system prompt remains unchanged between exercises, while the user prompt contains dynamic elements for each exercise, such as questions, correct answers, grading rubric, and examples. Except for student submission, all other entries are provided by the TA in advance. The response is a JSON structured text with predefined fields.}
    \label{fig:prompt-rubric-examples}
\end{figure*}

Student answers are evaluated using a single query to an LLM comprised of a generic system prompt and exercise-specific user prompt. While the system prompt contains general instructions and grading guidelines and remains the same for all exercises, the user prompt contains exercise-specific information, including the question, the predefined correct answer, the student submission, the grading rubric, and several TA-graded examples. The overall structure of our prompts is shown in Fig.~\ref{fig:prompt-rubric-examples}.
The prompts consist of two key components: the grading rubric, which provides structured grading criteria and point allotments for each exercise, and the grading examples of several TA-corrected student submissions of the particular exercise. We describe each of these in more detail below.

Each grading rubric consists of one or more grading criteria, each of which specifies a particular theme or aspect of the answer that must be included in the submission in order to satisfy the criteria (see blue highlighted text in Fig.~\ref{fig:prompt-rubric-examples}). Each criterion is allotted a certain number of points, which are tallied up into a final exercise score.
Each criterion can include an explanation section for providing informative feedback, but it is not used in grading. This is especially useful when the explanation is more involved and refers to aspects that students are not required to mention but are helpful to the explanation. The points from the satisfied criteria are tallied up into a final numeric score. In some instances, adding up the allotted points from all satisfied criteria would result in a score over 100\%. In these cases, we include an equation that specifies the exact computation of the final score (highlighted in red in Fig.~\ref{fig:prompt-rubric-examples}).

Fig.~\ref{fig:prompt-rubric-examples} shows one particular grading rubric comprised of three grading criteria. Each criterion is accompanied by an explanation. In this example, criteria A and B denote both parts of the correct answer (0.5 points each), but partial points can also be achieved via criteria C (0.25 points). Since a comprehensive student answer could satisfy all three criteria, simply adding the points together would yield a score of 1.25 points. Therefore, we include an expression in the preamble of the rubric table specifying how the final score should be obtained (see Fig.~\ref{fig:prompt-rubric-examples}, red).
Our grading criteria in this study are only additive, i.e., the points for each grading criteria match are added up. It is never subtractive, but we have no reason to suspect that that wouldn't work.

The grading examples section contains up to 10 examples of graded submissions. In the present study, graded examples are first grouped based on uniquely satisfying grading criteria (e.g., \textit{satisfies A but does not satisfy B}). Then, we uniformly sample a group and a sample within it, to obtain the most diverse range of graded examples assigned to the TA and TA-GPT-corrected groups.
%The implication here is that in the initial iteration of the course, some submissions need to be graded by humans, but these can be reused in future iterations.
%In order to keep the prompt as concise as possible and avoid issues with LLM context lengths, while still keeping our examples as representative and diverse as possible, we sample examples uniformly from each group. %Martin: kaj je tu group?

As shown in Fig.~\ref{fig:prompt-rubric-examples}, LLMs are prompted to return a structured response containing the submission score and written feedback, as well as a list of which of the satisfied rubric criteria. We have found that requiring LLMs to explicitly list the satisfied grading criteria improves results.
While we could programmatically parse the list of satisfied criteria and compute the total score of each submission, we have found that LLMs reliably handle this task and that mistakes are extremely rare, with only Llama-8B making a single mistake when tallying up the points out of 333 submissions (0.3\%), which we later corrected by hand.

%The task of LLMs is not to directly assess and grade each submission, but to determine which of the criteria the submission matches, and simply add up the allotted points. In this sense, the LLMs don't really behave like graders, but as classifiers that determine whether the submission fulfills some criteria.

\subsection{Preparing Grading Rubrics}
\label{sec:prep-grading-rubric}

We next describe our approach to preparing structured grading rubrics that are compatible with LLMs.
Initially, we prepare a preliminary version of the grading rubrics based on discussions among TAs and course instructors, specifying correct answers and which answers may deserve partial points.
We then manually correct a sample of student submissions and make adjustments to the rubrics as needed. As part of this study, 25\% of the text-based submissions are assigned to the TA or TA-GPT-revised grading groups, which we use for refining the grading rubrics.
To verify that the grading rubric is understandable to LLMs, we evaluate these same submissions using GPT-4o and manually inspect any mismatches between TA-assigned and LLM-assigned scores. In case of systematic differences in the LLM-assigned scores due to, e.g., a poorly worded prompt, we revise the grading rubric as needed.
We then perform a second evaluation using GPT-4o with the revised grading rubric and inspect whether any identified errors were resolved.

In practice, major changes to the grading rubric were rare. Most of our revisions involved rewording and clarifying ambiguous criteria.
Although we could theoretically continue refining the grading rubric until a perfect agreement between TA grades and GPT-4o is reached, we have found that one round of refinements is often enough to identify and correct majority of systematic errors. Additionally, further refinements of this kind would result in us overfitting our prompts to both our selected validation LLM, in our case GPT-4o, and the sample of manually corrected submissions.

\section{Results}

There are two aspects of grading that we consider here, both of which inform students about their performance: a numeric score assigned to each exercise and written feedback on their assignment.
In order for LLMs to make viable graders, they must be able to perform reasonably on both of these tasks. We here answer the following two questions.
Firstly, do the LLMs provide accurate grades? For LLMs to be effective in the classroom, they must grade student submissions accurately or at least align closely with the evaluations of human TAs.
The second component of grading is providing students with feedback on their submissions, highlighting errors and potential avenues of improvement.

\subsection{LLM Grading Accuracy}

Student submissions assigned to the TA and TA-GPT-revised grading groups were manually evaluated by human TAs. This subset accounts for 25\% of total student submissions, resulting in a training set of 670 manually evaluated submissions across 36 different text-based exercises. Each submission was assigned a score between 0 and 1, following the grading rubrics outlined in Sec.~\ref{sec:prompts}.
To elucidate the strengths and weaknesses of LLMs, we categorized each of the 36 exercises into five difficulty categories: ``trivial'' ($N=5$, $\mu_\text{score}=0.96$, $95\%$ CI $[0.92, 0.99]$), ``easy'' ($N=14$, $\mu_\text{score}=0.92$, $95\%$ CI $[0.89, 0.95]$), ``medium'' ($N=11$, $\mu_\text{score}=0.81$, $95\%$ CI $[0.75, 0.86]$), ``hard'' ($N=4$, $\mu_\text{score}=0.40$, $95\%$ CI $[0.31, 0.49]$), and ``open-ended'' ($N=2$, $\mu_\text{score}=0.90$, $95\%$ CI $[0.83, 0.96]$).
These manually assigned grades were then used as the gold standard for evaluating the performance of different LLMs. By comparing the LLM-generated scores to those assigned by human TA graders, we can quantitatively assess their accuracy and determine whether their performance is acceptable for classroom use.

At first glance, the most direct evaluation of LLM graders would be to compare the number of points assigned by LLMs to those assigned by human TAs. While intuitive, this approach would not, however, provide a reliable assessment of LLM performance in our specific scoring setup.
Numeric scores for each exercise are obtained by summing up the number of points allotted to each satisfied grading criterion in the grading rubric. For exercises with a single grading criterion, 100\% is awarded if the LLM correctly identifies that the submission satisfies the criterion. However, for exercises with multiple criteria, LLMs must correctly match all four criteria to award a full score. Consequently, treating the problem as a binary classification task is more appropriate, where LLMs determine whether a submission satisfies particular criteria.
While many different metrics are available for binary classification, we here report the classification accuracy (CA) as it directly relates to the proportion of correctly matched criteria and allows us to easily identify LLM grading biases in terms of leniency and strictness. Each of the 36 questions is comprised of 1 to 4 grading criteria, resulting in a total of 61 grading criteria segments.

\begin{figure*}[htp]
    \centering
    \includegraphics[width=\linewidth]{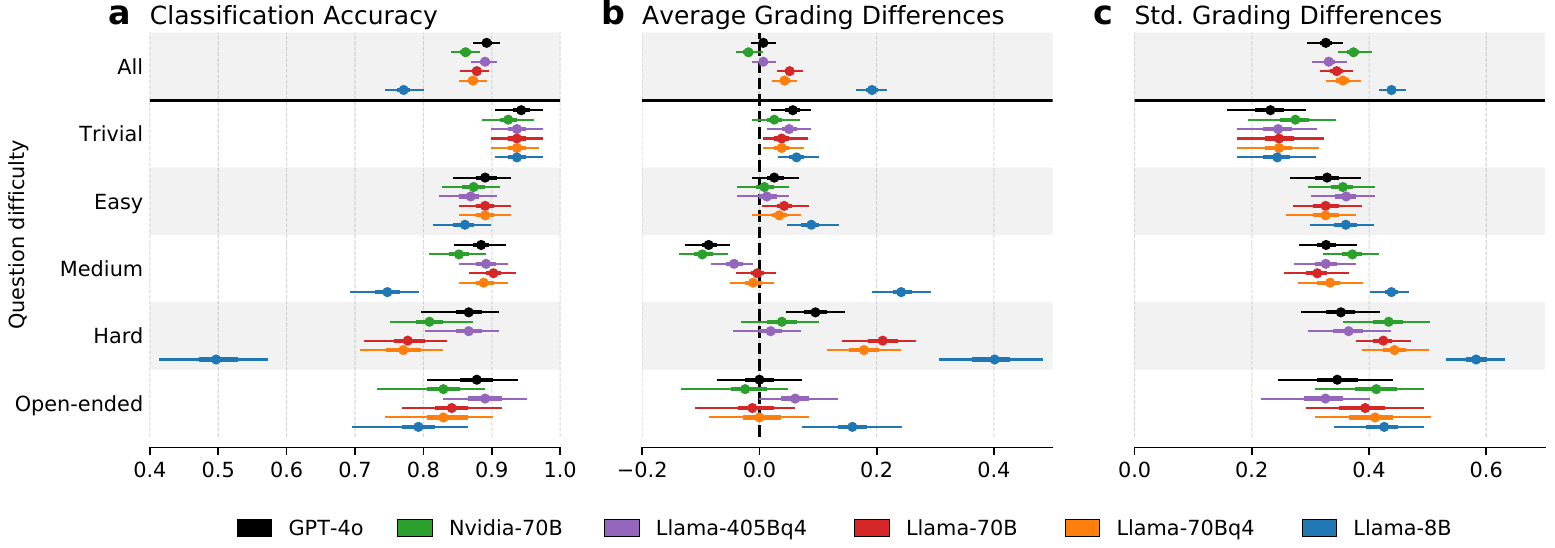}
    \caption{LLM performance on predicting grading criteria. TA grades represent the gold standard. 95\% credible intervals (CI) of summary statistics are obtained using bootstrap samples. \textbf{a}) Classification accuracy of LLMs predicting each satisfied criteria as a binary classification. \textbf{b}) The average grading difference in prediction indicating more lenient (positive) or stringent (negative) grading by the LLM compared to TAs. \textbf{c}) The standard deviation of grading difference indicates consistency among models.}
    \label{fig:llm-performance}
\end{figure*}

The top panel of Fig.~\ref{fig:llm-performance}.a shows the overall classification accuracy of each of the LLMs. Overall, LLMs achieve strong performance, with average CA scores ranging between 85\% and 90\%. One notable exception is Llama-8B, which achieves a relatively poor CA of 75\%.
When grouping exercises by difficulty, we notice a decrease in CA as the difficulty of the exercises increases. This is likely because harder-to-answer questions often receive wildly varying answers that are impossible to foresee and define their scoring within the prompts. Hard questions, in particular, often require longer answers that sometimes contain mathematical equations, which may be difficult for models to categorize appropriately.
One particularly interesting category of questions is open-ended questions, which pose an interesting challenge. For open-ended questions, it is often impractical to exhaustively list all possible correct, and the final judgment must often be made by the LLM. Despite this, LLMs generally achieve solid performance, achieving accuracies between 80\% and 90\%.
One interesting observation here is that in the ``hard'' and ``open-ended'' categories, model performance appears to closely match the number of model parameters. Both GPT-4o and Llama-405Bq4 achieve similar performance, while the 70B models all achieve slightly lower performance. Llama-8B performs substantially worse still.

While the classification accuracy conveys the number of correctly identified criteria, it does not reveal whether models tend to be more lenient or stringent than TAs. In Fig.~\ref{fig:llm-performance}.b, we plot the average differences in the matched grading criteria. Positive values indicate that LLMs were more generous, classifying more criteria as satisfied than TAs. Negative values indicate that models failed to report many of the criteria that TAs marked as satisfied, resulting in lower final grades. For trivial and easy questions, the models exhibit little systematic bias with their means close to zero. For medium-difficulty questions, the models appear to show low levels of negative bias, while the opposite is true for difficult questions, where most models are significantly more generous than TAs. In open-ended questions, the majority of models exhibited no systematic bias.
Finally, Fig.~\ref{fig:llm-performance}.c shows the variance of this bias. The majority of models appear to be equally consistent with one another.

The obvious exception to the above is Llama-8B, which exhibits poor performance across the board and is overly generous in its grading, e.g., for hard questions, Llama-8B over graded about 45\% of submissions and under graded about 5\% of submissions, correctly grading only 50\% of submissions.
The poor performance of Llama-8B across all performance metrics indicates its unsuitability for its use as an assignment grader in the classroom. Its poor performance could be due to several factors.
Firstly, our prompts are quite long, and perhaps Llama-8B struggles with the context size. Interestingly, however, this does not appear to be the case with easier questions, casting doubt on this explanation.
Secondly, our prompts were not tailored to any one specific model, and we only used GPT-4o in our dry run. Perhaps fine-tuning the prompts would allow us to achieve better performance on Llama-8B. However, given that the remaining models did not encounter such difficulties, we anticipate that the most likely explanation is due to a final possible explanation -- the inherent limitations of the smaller model.
This is corroborated by the fact that Llama-8B appears to have the most difficulties with harder questions which typically involve more involved questions. These questions typically result in longer, more nuanced answers, for which Llama-8B perhaps lacks the reasoning capabilities to adequately parse and match to the grading criteria. On the other hand, 4-bit quantized Llama-70Bq4 with similar hardware requirements performed much better and achieved near non-quantized performance.

From the analysis above, we make the following observations:
\begin{enumerate}
    \item With the exception of Llama-8B, all models achieve adequate performance, demonstrating high accuracy when determining whether a particular submission satisfies predefined grading criteria and exhibits low levels of systematic bias.
    \item Model performance generally correlates with their number of parameters. The larger GPT-4o and Llama-405Bq4 models perform favorably to the 70B parameter models, which in turn outperform the 8B parameter Llama variant.
    \item Quantization appears to have negligible effect on performance, as the quantized variant of Llama-70B model achieves comparable performance to its full-precision counterpart.
    \item Although none of the models achieve perfect accuracy, we have determined their margin of error to be acceptable. Given the general direction of the grading biases, we anticipate little student pushback. Furthermore, students who suspect grading errors can request a manual review.
\end{enumerate}

\subsection{Impact of Including Grading Rubric and Grading Examples}

\begin{figure*}[htp]
    \centering
    \includegraphics[width=\linewidth]{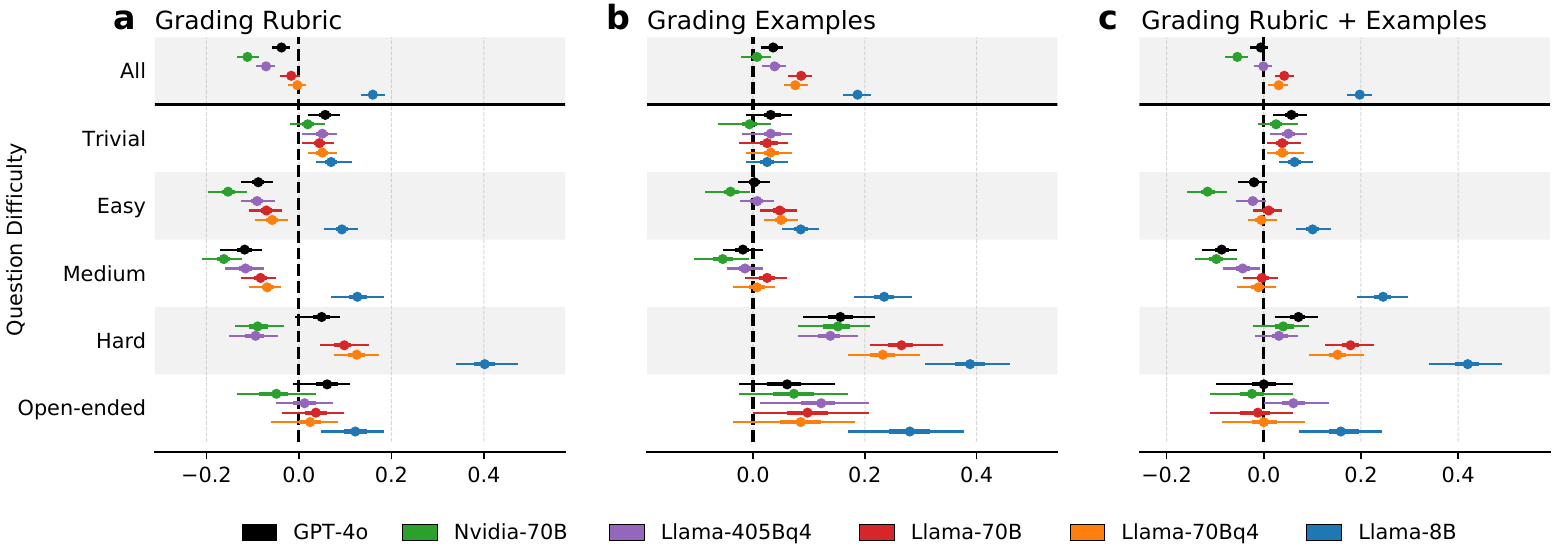}
    \caption{Importance of grading rubric and graded-examples on LLM performance. The scale relates to systematic bias with respect to TA grades. 95\% credible intervals (CI) of summary statistics are obtained using bootstrap samples. \textbf{a}) LLM performance using only the TA-defined grading rubric in the user prompt. \textbf{b}) LLM performance using only TA-graded examples without the grading rubric. \textbf{c}) LLM performance using both a grading rubric and graded examples in the user prompt. Both rubric and examples are used in production.}
    \label{fig:prompt-performance}
\end{figure*}

In the previous section, we examined the performance of different LLMs using prompts that included both grading rubrics and grading examples. Here, we investigate the effects of excluding each of these elements from the prompt. In Fig.~\ref{fig:prompt-performance}, we report the mean differences in the matched grading criteria for the three different prompt variants.

The top rows of Fig.~\ref{fig:prompt-performance} show the performance of the six LLMs across the three prompt variants.
Prompts that included only grading rubrics led to stricter grading, with LLMs less likely to match grading criteria (Fig.~\ref{fig:prompt-performance}.a). On the other hand, prompts that included only grading examples resulted in more lenient grading, as LLMs were overly generous (Fig.~\ref{fig:prompt-performance}.b). Including both the grading rubric and grading examples produced the best results, achieving a middle ground between the two individual results. Curiously, these biases did not greatly affect their classification accuracy, which was predominantly not statistically significant; all three variants achieved similar CA scores (not shown for brevity).

We might expect that, given enough examples, LLMs could infer the grading rubric internally, potentially eliminating the need for course instructors and TAs to prepare detailed grading rubrics.
The results from Fig.~\ref{fig:prompt-performance}.b indicate that, for simpler questions, LLMs achieve satisfactory performance using grading examples alone. However, for harder and open-ended questions, we observe a marked drop in performance.
We hypothesize that this may be due to the increased variability in student responses. Simpler questions tend to have more straightforward answers with limited variation. Since we include several manually graded student submissions in the grading prompt, most student responses will likely be similar to the grading examples, giving LLMs a blueprint for the desired response.
In contrast, answers to more difficult and open-ended questions are often longer and more varied, making it less likely that the grading examples will cover the wide range of possible answers. For these more difficult questions, providing a grading rubric is essential (see Fig.~\ref{fig:prompt-performance}.c).

\subsection{Student Preferences for LLM-based Feedback}

Feedback is considered a fundamental aspect of the learning process, and effective feedback has been shown to improve learning outcomes~\cite{Paterson2020}.
To assess whether students preferred human-written feedback or feedback generated by LLMs, we asked students to rate the feedback received for each text-based answer after receiving feedback for each assignment.
We received student satisfaction scores for a total of 1,527 answers, of which 1,189 answers were correct, and 338 answers were incorrect or partially correct.

We determine student preferences for individual grader feedback using a Bayesian mixed-effects linear regression model~\cite{Kruschke2015}:
\begin{align}
\mu_i &= \gamma_{m_i} + \eta_{e_i} + \psi_{s_i} + \alpha \cdot \text{score}_i + \tau \cdot \text{total}_i, \label{eq:model} \\
y_i &\sim \text{OrderedProbit}(\mu_i, \text{cutpoints}), \notag
\end{align}
where $y_i$ denotes the student satisfaction score for a particular text-answer $i$. $\gamma_{m_i}$ denotes the factor for the assigned grader $m_i$, $\eta_{e_i}$ represents the factor assigned to each exercise, accounting for different difficulties of exercises and $\psi{s_i}$ accounts for the different student biases. Students who received higher grades on a particular exercise and on the assignment as a whole are more likely to assign a high satisfaction rating. We account for these effects using coefficients $\alpha$ and $\tau$, modeling the effects of exercise scores and total assignment scores, respectively.
We assign an uninformative prior $\mathcal{N}(0, 2)$ on all model parameters.
Inference was performed using the Stan library using Hamiltonian Monte Carlo sampling (HMC)~\cite{StanLang}.

\begin{figure*}[htp]
    \centering
    \includegraphics[width=\linewidth]{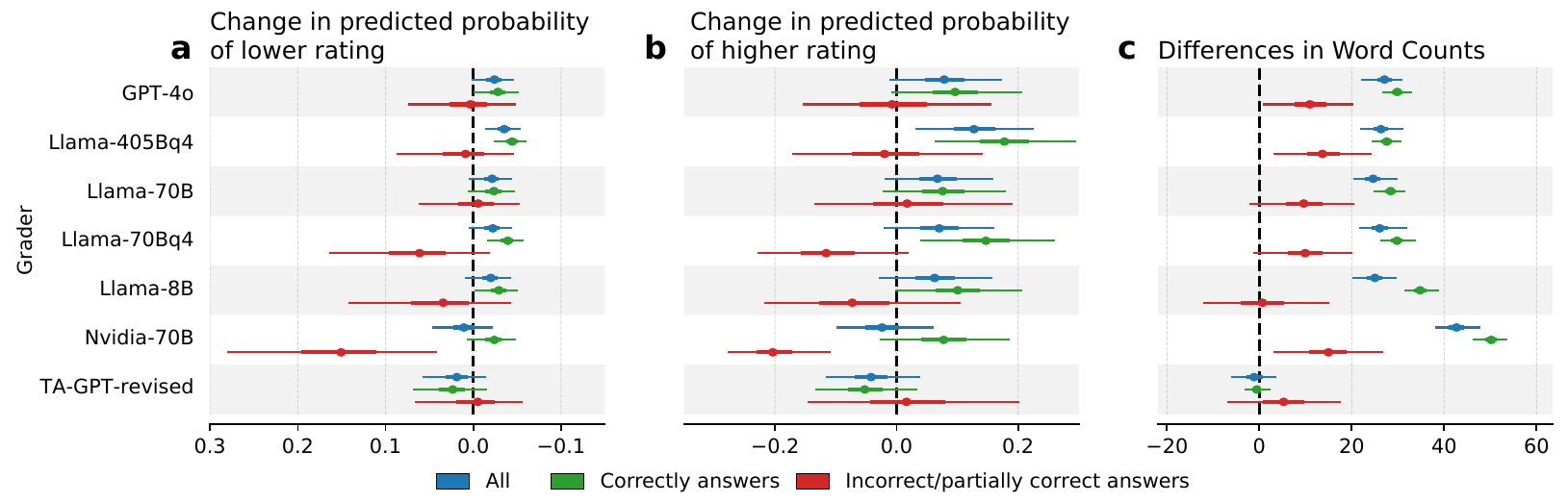}
    \caption{Group factor in relation to student preference. Due to the correlation between grading group sample values, the TA grading group is used as a reference, and the values show an increase and a decrease in satisfaction with TA written feedback.}
    \label{fig:student-preferences}
\end{figure*}

We visualize the student preferences of the different graders in Figs.~\ref{fig:student-preferences}.a and b. Due to the inherent correlation between group factors during HMC sampling, we report the differences in satisfaction between each of the grading groups and TA feedback. Figs.~\ref{fig:student-preferences}.a and b show the average differences in probability that the student would assign a higher satisfaction score and a lower satisfaction score if that a particular grader was used instead of receiving feedback written by TAs.

With the exception of Llama-405Bq4, which students appear to prefer slightly, the results indicate that, overall, students exhibit no significant preferences for any of the graders.
% student expectations are met when
%An interesting hypothesis is that perhaps student preferences change when
However, when examining feedback preferences separately for correctly and incorrectly answer questions, an interesting difference begins to appear.
We the effects of grader preferences on correctly vs incorrectly graded answers, we use the same model as in Eqn.~\eqref{eq:model}, but use two sets of grader factor parameters, one for correctly and one for incorrectly answered responses. We connect each pair of grader factors using a Gaussian hyperprior.
Figs.~\ref{fig:student-preferences}.a shows that students showed no significant dislike for feedback generated by any of the LLMs. One notable exception to this is Nvidia-70B, for which students were around 15\% more likely to respond negatively than if they were graded by a human TA. This suggest that, with the exception of Nvidia-70B, the written feedback generated by LLMs is generally as good as feedback written by human TAs.
Figs.~\ref{fig:student-preferences}.b shows the change in predicted probabilities for higher ratings, i.e., how much more likely are students to rate a piece of feedback higher if it were written by an LLM. Interestingly, although the effect size is small, students seem to prefer feedback written by LLMs over feedback written by human TAs. This holds particularly true for feedback to correctly answered questions. For feedback on incorrectly answered questions, TA feedback achieves similar satisfaction levels to LLM-written feedback.

One possible explanation for this effect is that when the answer is correct, human TAs tend to respond with short feedback, indicating only that the answer was correct. Examples of this feedback might include ``ok'' or ``That's right.''. On the other hand, LLMs tend to generate longer feedback, elaborating why the particular answer was deemed correct. Fig.~\ref{fig:student-preferences}.c shows the differences in average feedback lengths between LLM graders and human TAs. The green lines, corresponding to correctly answered questions, tend to be much lower for TA written feedback.
However, when the answer is incorrect or partially correct, the feedback from TAs tends to focus on the missing or incorrect aspects of the answer, highlighting what was wrong and explaining the correct solution, thus producing longer feedback.
%Interestingly, asking ChatGPT to revise the feedback didn't result to have much impact on the feedback length and the changes are often superficial. For instance, a common revision might replace the human written feedback ``That's right.'' with ``Nicely done!''.

\subsection{Student Attitudes Towards LLM-based Grading}

\begin{figure}[htp]
    \centering
    \includegraphics[width=0.75\linewidth]{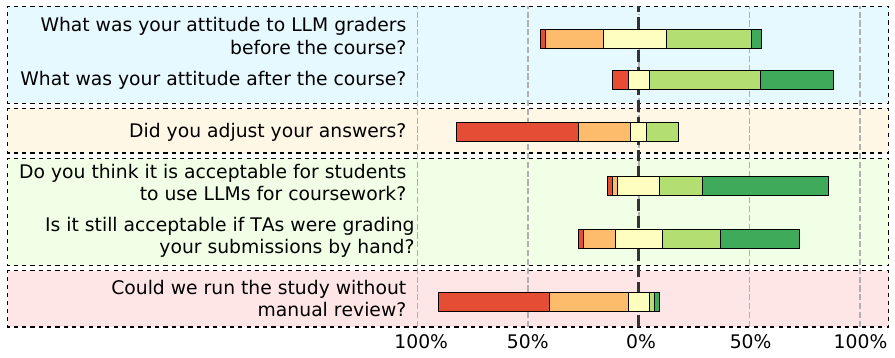}
    \caption{Results of the final survey. Questions are asked on a five-point Likert scale. Red bar colors correspond to negative attitudes and disagreement, while green bar colors indicate positive attitudes and agreement}
    \label{fig:final-survey}
\end{figure}

At the end of the semester, we presented the preliminary findings of this study to the students in the classroom. Following the session, students were asked to complete a short, anonymous survey regarding their attitudes toward the use of LLMs as assignment graders and whether their attitudes had changed over the course of the semester. A total of 42 students responded to the survey. Selected results are shown in Fig.~\ref{fig:final-survey}.

We first asked students whether they felt it was appropriate for us to grade their assignments using LLMs before the beginning of the course. Student responses were mixed, with an average score of $3.2$. Encouragingly, after completing the course, students were much more open to LLM graders, with the average score increasing to over $4.0$.

We also asked students whether they had used any LLM-enabled tools while working on the assignments. Over 92\% of students reported using such tools, with 90\% using ChatGPT for solving the assignment and 46\% using Copilot for code generation. One student also reported using ChatGPT to better understand the assignment instructions. Most students expressed that it is fair for them to use LLMs when solving the assignments if graded by LLMS ($\mu=4.2$), but feel more hesitant about it when graded by human TAs ($\mu=3.8$)(Fig.~\ref{fig:final-survey}). Although students knew their answers might be graded by an LLM, they largely reported on keeping their answering style.

In the present study, students could request a manual review of their grades at any time. 
Consistent with prior research~\cite{Chiang2024}, students strongly felt it would be unacceptable not to have the option to request a manual review.

\section{Recommendations and Guidelines}

Based on our semester-long experience and the results of our analysis, we offer the following recommendations and guidelines for incorporating LLMs into assignment grading workflows:
\begin{enumerate}
    \item \textbf{Use Structured Grading Rubrics:} Develop structured grading rubrics and include specific sections for explanations. This enables LLMs to provide clearer feedback, particularly for more difficult questions.
    
    \item \textbf{Include Graded Examples:} Include graded examples of the student submissions. These examples help LLMs better understand TA grading style and expectations.

    \item \textbf{Test New Grading Rubrics:} When preparing grading rubrics, conduct a dry run on a sample of manually graded student submissions to identify potential systematic grading errors. Pay close attention to the wording of criteria, as LLMs may sometimes be unpredictably pedantic, and small changes in wording can significantly impact grading accuracy. Any refinements should further be validated, ideally on a new sample of student submissions, to avoid overfitting.

    \item \textbf{Open-Source LLMs:} If selecting an open-source LLM, we recommend selecting the largest LLM your hardware can support. Quantization appears to have negligible effects on an LLM's grading capabilities compared to their full-precision counterparts, so prioritize larger quantized models over smaller full-precision models. In terms of grading performance, open-source LLMs perform as well as their commercial counterpart.s

    \item \textbf{Allow Requesting Manual Review:} Provide students with the option to request a manual review of their grades, as LLMs still make occasional errors.
\end{enumerate}

\section{Conclusion}

We presented a study on the use of LLMs for grading written assignments in the Introduction to Bioinformatics course during the 2024/25 academic year. By implementing and evaluating LLM graders in a real-world classroom setting, we found that automated grading can achieve performance comparable to human teaching assistants in both scoring and feedback generation. Our findings show that well-designed grading rubrics and examples graded by teaching assistants help make automated grading work well in courses with many students.

Quality feedback improves student learning and academic achievement, but providing timely, clear, and personalized feedback is challenging in large university courses~\cite{Paterson2020}. Impey \etal~\cite{Impey2024} showed that LLM graders outperform peer grading on essay-type answers when provided with TA-generated answers and rubrics. Our results confirm these findings and further demonstrate that incorporating TA-graded examples enhances the quality of machine-generated feedback, making it even more aligned with student preferences.

Our results show that open-source models perform on par with commercial alternatives. For example, Llama-405Bq4 achieved comparable results to GPT-4o across all evaluated criteria. This suggests that, with sufficient hardware resources, universities could deploy their own instances of LLM graders without compromising performance. Such an approach could also alleviate the substantial financial costs associated with commercial solutions, as highlighted by Chiang \etal~\cite{Chiang2024}. While the comparable performance of open-source models is promising, their high hardware demands may pose challenges for many university departments. Recent research has focused on developing smaller models that can achieve similar performance to larger ones~\cite{team2024gemma}, and it is envisioned that, in the future, grading could be performed locally on consumer-grade laptops, making it accessible to everyone. However, this capability is not yet a reality. An alternative approach could involve fine-tuning existing models to enhance performance, as studies have shown that even small amounts of domain-specific data can lead to significant improvements~\cite{Katuka2024}.

Our study has several limitations. First, due to their probabilistic nature, LLMs can generate different grading responses when queried multiple times, even with identical prompts. While many providers and implementations offer options to achieve more consistent results by adjusting the temperature parameter, some variability remains~\cite{Jauhiainen2024}. Although we did not explicitly investigate the impact of temperature in this study, we set the temperature of all LLMs to zero to minimize randomness. Second, previous studies have reported instances of students engaging in prompt-hacking—where submissions contain deceptive instructions, such as directing the LLM to assign the maximum possible score~\cite{Chiang2023}. To mitigate this, we incorporated anti-cheating measures into our system prompts; however, we did not observe any prompt-hacking attempts throughout the semester. While we did not explicitly prohibit this behavior, students may have refrained from such practices, knowing that their submissions could be reviewed by human teaching assistants. In an LLM-only grading environment, students might be more inclined to exploit such vulnerabilities. Therefore, implementing robust safeguards to detect and prevent malicious input remains essential.

Our study introduces an innovative approach to automated grading by conducting a real-classroom evaluation in the Introduction to Bioinformatics course, a carefully designed program previously reported at ISMB 2024~\cite{Policar2024}. With a large number of students participating in a randomized study, we systematically compared the performance of multiple open-source and commercial LLMs. Our findings demonstrate that open-source models can achieve comparable results to commercial alternatives, offering institutions greater control over their grading processes. These contributions provide valuable insights for the broader adoption of LLM-based grading in bioinformatics education and beyond.

%In ~\cite{Impey2024} study, the authors found that LLMs could also be used to generate grading rubrics of similar quality to human-designed criteria.

\section{Competing interests}
No competing interest is declared.

\section{Author contributions statement}

P.G.P., M.\v{S}., and B.Z. designed the study. P.G.P. and M.\v{S}. carried out the study and analysis. P.G.P. and M.\v{S}. wrote the paper. B.Z. and T.C. reviewed and edited the paper.

\section{Acknowledgments}
We thank Jaka Koko\v{s}ar for assisting with running the LLMs locally, Janez Dem\v{s}ar for valuable input on study design, Erik \v{S}trumbelj for statistical analysis recommendations, and the students of the Introduction to Bioinformatics course at the University of Ljubljana (2024/25) for their participation in the study and useful feedbacks.

\section{Funding}
This work is supported by Research Program (P2-0209) and Young Researcher (57111 to M.Š.) grants from Slovenian Research
and Innovation Agency.

\bibliographystyle{unsrt}  

%\bibliographystyle{plain}
\bibliography{references}

\end{document}